\documentclass[runningheads]{llncs}
\usepackage{graphicx}
\usepackage{subcaption}
\usepackage{amsmath,amssymb}
\usepackage{multirow}
\usepackage{hyperref}

\usepackage[utf8]{inputenc}

\DeclareMathOperator{\E}{\mathbb{E}}

\begin{document}
\title{Context-encoding Variational Autoencoder for Unsupervised Anomaly Detection}
\titlerunning{ceVAE for Unsupervised Anomaly Detection}
\author{David Zimmerer\inst{1} \and
Simon A.A. Kohl\inst{1} \and
Jens Petersen\inst{1}\and
Fabian Isensee\inst{1}\and
Klaus  H. Maier-Hein\inst{1}}
\authorrunning{D. Zimmerer et al.}
\institute{Division of Medical Image Computing, German Cancer Research Center (DKFZ),
Heidelberg, Germany}
\maketitle              %
\begin{abstract}
Unsupervised learning can leverage large-scale data sources without the need for annotations. In this context, deep learning-based auto encoders have shown great potential in detecting anomalies in medical images. However, state-of-the-art anomaly scores are still based on the reconstruction error, which lacks in two essential parts: it ignores the model-internal representation employed for reconstruction, and it lacks formal assertions and comparability between samples. We address these shortcomings by proposing the Context-encoding Variational Autoencoder (ceVAE) which combines re\-con\-struction- with density-based anomaly scoring. This improves the sample- as well as pixel-wise results. In our experiments on the BraTS-2017 and ISLES-2015 segmentation benchmarks the ceVAE achieves unsupervised ROC-AUCs of 0.95 and 0.89, respectively, thus outperforming state-of-the-art methods by a considerable margin. 

\end{abstract}

\section{Introduction}

The ongoing technological advancements in medical imaging result in an ever-increasing image quality and quantity in clinical, scientific and industrial settings leading to an increasing amount of conditions that become detectable \cite{vernooij_incidental_2007}. 
Currently, the inspection of most medical image data is performed manually by trained physicians, which is time and resource consuming and does not scale very well. Furthermore, while medical experts have a high sensitivity to the specific condition in question, they are vulnerable to inattentional blindness, leading to high miss-rates of unexpected anomalies and conditions \cite{drew_invisible_2013}.
Missing an (incidental) finding can have grave consequences for the patient and prevent the early detection of relevant medical conditions
\cite{bluemke_chapter_2012}.
Machine-learning based support systems might be able to alleviate this problem, but usually require a large annotated dataset for every condition and modality.
This is a major drawback that currently hampers the application of machine learning in clinical practice. 
Also, this approach still fails on conditions not explicitly represented in the training database.
Anomaly detection aims at identifying unexpected, abnormal data points given a set of normal data samples only, thus highlighting interesting regions for further manual inspection.
Importantly, it does not require supervision in form of manual annotations, is independent of human judgment errors and, instead, automatically internalizes the appearance of normal tissue to recognize anomalies.
\vspace{-1.5em}

\subsubsection{Contribution} 
In this paper, we present a novel anomaly detection method that can be used to identify and localize abnormal regions in medical images. Our contributions are (i) we show how to combine a Context Encoder \cite{pathak_context_2016} with a Variational Autoencoder \cite{kingma_auto-encoding_2013,rezende_stochastic_2014} to improve anomaly scores, (ii) to the best of our knowledge we are the first to include the deviations (KL-divergence) of the posterior from the prior of the latent variable distributions in a Variational Autoencoder for pixel-wise anomaly localization, (iii) we fuse the deviations from the prior in a Variational Autoencoder with the reconstruction error to improve the localization, (iv) with this approach we are able to outperform the state-of-the-art unsupervised approaches on 
two public segmentation challenges \cite{chen_deep_2018}.

\vspace{-0.7em}
\section{Related Work}
\vspace{-0.5em}

\subsection{Autoencoders}
\vspace{-0.5em}

An Autoencoder (AE) is trained to reconstruct its input $x$ from a learned representation $z$ \cite{ballard_modular_1987}. It consists of two parts, an encoder $z = enc(x)$,  which encodes the input $x$ to a learned feature representation $z$, and a decoder $\hat{x} = dec(z)$ which attempts to recapture the original input by decoding the representation. %
For a deep convolutional AE the encoder $enc_{\theta}(x)$ and decoder $dec_{\gamma}(z)$ each are modeled as deep-convolutional networks with parameters $\theta$ and $\gamma $, respectively. 
Hence, the training of a deep AE can be formalized as:
\vspace{-0.4em}
\begin{equation}
\label{eq:ae}
\min_{\theta, \gamma} \sum_x L_{\textit{rec}} (x, \hat{x}) \textnormal{, with } \hat{x}=dec_{\gamma}(enc_{\theta}(x)) .
\vspace{-0.8em}
\end{equation}
A common choice for the reconstruction error $L_{\textit{rec}} (x, \hat{x})$ is the mean-squared error (MSE): $L_{\textit{rec}} (x, \hat{x})  = L_{\textit{MSE}} (x, \hat{x}) = || x - \hat{x} || ^2 .$
To reconstruct the image truthfully, the encoder has to encode the information of the input into the feature vector $z$. To learn more suitable representations $z$ different variations of AE have been proposed \cite{rifai_contractive_2011,vincent_stacked_2010}: 
\vspace{-1.0em}

\subsubsection{Denoising Autoencoder}
A Denosing Autoencoder (DAE) is trained to reconstruct the unperturbed data sample $x$ from an input sample that has been subjected to noise. This results in more robust and perturbation invariant representations \cite{vincent_stacked_2010}. The most commonly used noise is additive Gaussian noise, i.e.:
$\tilde{x} = x + \epsilon \textnormal{ , with } \epsilon \sim \mathcal{N}(0, \sigma^2) , $
for a small value $\sigma$. Thus $\hat{x}$ in Eq. (\ref{eq:ae}) becomes $\hat{x}=dec_{\gamma}(enc_{\theta}(\tilde{x}))$.
Context Encoders (CEs) are a special class of DAEs where instead of the commonly used additive Gaussian noise local patches of the input are masked out. This can be interpreted as a variation of salt-and-pepper noise and was shown to results in better generalizing representations, which in addition to appearance also captures semantic information of the input \cite{pathak_context_2016}. 
\vspace{-1.0em}

\subsubsection{Variational Autoencoders}
\label{sssec:VAE}
A Variational Autoencoder (VAE) \cite{kingma_auto-encoding_2013,rezende_stochastic_2014} assumes a latent variable model where a latent variable $z$ causes the observation $x$, facilitating a lower bound of the probability of a data sample with 
\vspace{-0.6em}
\begin{equation}
\label{eq:elbo}
\log p(x) \geq -D_{KL}(q(z|x) || p(z)) + \E_{q(z|x)} [\log p(x|z)] ,
\vspace{-0.6em}
\end{equation}
which is often termed the Evidence Lower Bound (ELBO). 
Here $p(z)$ is the prior distribution of the latent variable, $q(z|x)$ is the approximate inference model and $p(x | z)$ is the generative model.  
By maximizing the ELBO, the probability distribution approximates the true data distribution and enables a probability estimate for a data sample. 
VAEs parameterize $q(z|x)$ and $p(x | z)$ by neural networks and for $p(z)$ and $p(x|z)$ assume diagonal Gaussian distributions:
\vspace{-0.4em}
\begin{equation}
\begin{split}
& q(z|x) = \mathcal{N}(z; f_{\mu, \theta_1}(x), f_{\sigma, \theta_2}(x)^2 ),\\
& p(x | z)=\mathcal{N}(x ; g_{\mu, \gamma}(z), c(z)^2), 
\vspace{-1.5em}
\end{split}
\end{equation}
where $f_{\mu}$, $f_{\sigma}$, and $g_{\mu}$ are neural networks with parameters $\theta_{1}$, $\theta_{2}$ and $\gamma$ respectively and $c$ is often chosen as constant. In analogy to AEs $f$ is called the encoder and $g$ is called the decoder.
The often used formulation for VAE training is:
\vspace{-0.4em}
\begin{equation}
\label{eq:vae}
\min_{\theta_1, \theta_2, \gamma} \sum_x D_{KL}( \mathcal{N}(f_{\mu, \theta_1}(x) , f_{\sigma, \theta_2}(x)^2 )) || \mathcal{N}(0, 1) ) + L_{rec}(x, g_{\mu, \gamma}(\tilde{z})), 
\vspace{-0.8em}
\end{equation}
with $\tilde{z}$ being sampled from $\mathcal{N}(f_{\mu, \theta_1}(x) , f_{\sigma, \theta_2}(x)^2 )$ using the reparametrization trick \cite{kingma_auto-encoding_2013,rezende_stochastic_2014} and MSE is chosen for $L_{rec}$.

\subsection{Anomaly detection}

\subsubsection{Classification-based methods}
are one class of unsupervised anomaly detection methods. A prominent example of the classification-based methods is the  One-Class Support Vector Machine (OC-SVM) \cite{scholkopf_estimating_2001}. The OC-SVM finds a decision boundary between the data features and the origin in features space to differentiate normal data from abnormal data.
\vspace{-1.2em}

\subsubsection{Reconstruction-based methods}
\label{ssec:recbased}
aim at truthfully reconstructing normal data samples while producing high reconstruction errors for abnormal data. As compared to Principal Component Analysis (PCA)-based reconstruction methods \cite{shyu_novel_2003}, %
AE-based reconstruction methods can better handle non-linear relations in the data. 
Reconstruction-based approaches are used in medical imaging almost exclusively, since they allow a pixel-wise anomaly detection and can delineate the pathological conditions.
Schlegl et al. \cite{schlegl_unsupervised_2017} use a generative adversarial network (GAN)-based method to estimate an anomaly score. 
Based on the assumption that a fully trained GAN can only produce samples from the learned data distribution, they use an iterative back-propagation algorithm to find the closest match to the sample of interest that the trained GAN can produce. The anomaly score is then derived from the similarity of the real and generated sample. 
Different AEs have been employed for anomaly detection in brain images. 
Baur et al. \cite{baur_deep_2018} employed VAEs and used the reconstruction error for localization of MS lesions on an in-house MRI dataset. In their experiments, the VAE with an adversarial reconstruction loss slightly outperformed a vanilla VAE. Chen et al. \cite{chen_unsupervised_2018,chen_deep_2018} show that a combination of a VAE with an adversarial loss on the latent variables can boost performance in detecting brain tumors in the BraTS 2015 MRI dataset 
\cite{bakas_advancing_2017}
using a pixel-wise reconstruction error. Pawlowski et al. \cite{pawlowski_unsupervised_2018} train different AEs on an in-house brain CT dataset with intracranial hemorrhages and traumatic brain injuries. Similar to the studies above, they consider the pixel-wise reconstruction error of different AE models for pixel-wise anomaly detection. In their evaluation, an AE with dropout sampling in the bottleneck layer slightly outperforms the other models.
Despite their frequent use most reconstruction-based methods have no formal assertions regarding the reconstruction-error, complicating the interpretation and the comparability of anomaly scores.
A more theoretically grounded improvement is given by Alain et al \cite{alain_what_2014}, showing that the denoising task in DAEs can lead to reconstruction errors that approximate the local derivative of the log-density with respect to the input. Consequently, the global reconstruction error for a whole sample reflects the norm of the derivative of the log-density with respect to the input. While this ``direction to normality''  can yield important clues, it is still not the probability of the data sample itself, still posing challenges for a sample-wise comparable and well-calibrated anomaly score.
Density-based models offer a solution for this problem.

\vspace{-0.8em}
\subsubsection{Density-based methods}
give a probability estimate for each data sample, allowing for a straight-forward normality-scoring and -ordering. 
This class can be further split into parametric and non-parametric algorithms and other methods. The non-parametric approaches, such as neighborhood-based methods
and clustering-based methods
estimate the data density locally and assign an anomaly score based on the probability of a new data sample  \cite{goldstein_comparative_2016}. Parametric approaches
assume a data distribution and fit the distribution parameters to the data. Due to the ``curse of dimensionality" 
\cite{bach_breaking_2017}
these methods, similar to the OC-SVM and PCA, often struggle in high dimensional data settings \cite{goldstein_comparative_2016}. 
VAEs \cite{kingma_auto-encoding_2013,rezende_stochastic_2014} are able to alleviate this problem \cite{bach_breaking_2017}, allowing to estimate abnormality scores on the basis of the evidence lower bound for a data sample 
\cite{kiran_overview_2018}.
Current anomaly detection methods in the literature however, employ VAEs for reconstruction \cite{an_variational_2015,baur_deep_2018,chen_unsupervised_2018,pawlowski_unsupervised_2018} and still use only the reconstruction-error model for anomaly scoreing, thus ignoring an essential part of the model. 
Moreover, to our knowledge, density-based approaches have not been explicitly applied to medical imaging, presumably since they do not directly give an anomaly score on a pixel level.

\vspace{-0.8em}
\subsubsection{Problem Statement}

Most medical imaging anomaly detection methods are based on the reconstruction error, mostly employing AE-variants for reconstruction.
However, for AE-based models, the reconstruction error lacks in two essential parts.
First, only considering the reconstruction error ignores all model-internal variations such as deviations of the latent representations from their normal ranges, which can indicate an anomaly, especially in case of a perfect reconstruction.
Second, the reconstruction error on its own
has in most cases no formal assertion and no theory-backed validity, rendering it unsuited as a well-calibrated and comparable anomaly score.

\vspace{-0.4em}
\section{Methods}
\label{sssec:ceVAE}

\begin{figure}[bt]
  \centering
    \includegraphics[width=0.75\textwidth]{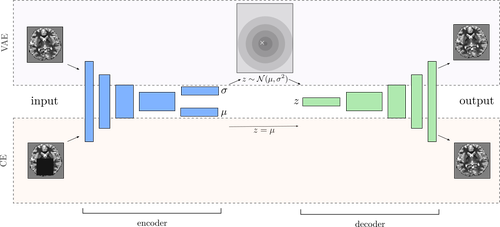}
  \caption{ceVAE model structure.}
\label{img:network}
\vspace{-1.0em}
\end{figure}

\vspace{-0.4em}
To alleviate the mentioned shortcomings, we present a novel anomaly detection method: Context-encoding Variational Autoencoder (ceVAE). By combining CE and VAE, we strive to use the model-internal latent representation deviations and a more expressive reconstruction error for anomaly detection on a sample as well as pixel level.
We define the ceVAE with fully convolutional encoders $f_{\mu}$, $f_{\sigma}$ and a decoder $g$, where the CE only uses the mean encoder $f_{\mu}$ to encode a data sample (with $f_{\mu}$ and $f_{\sigma}$ sharing most of their weights \cite{kingma_auto-encoding_2013}, see Fig \ref{img:network}).

\vspace{-0.4em}
\subsubsection{CE branch} We subject a sample $x$ to context encoding noise by masking certain regions in the input (randomly sized and positioned). The CE branch is trained by reconstructing the perturbed input $\tilde{x}$ using $f_{\mu}$ as encoder and $g$ as decoder: $L_{rec_{CE}}(x, g(f_{\mu}(\tilde{x})))$.
As described in Sec. \ref{ssec:recbased}, the denoising task is expected to gear the reconstruction error towards the approximation of the derivative of the log-density with respect to the input $\frac{\partial \log p(x)}{\partial x}$.
This, on its own, could be helpful for detecting anomalous parts in a data sample since it can yield better calibrated and interpretable reconstruction errors \cite{alain_what_2014}.
At the same time, CEs have been shown to result in more discriminative, semantically richer representations \cite{pathak_context_2016}.
This is expected to have a positive influence on the expressiveness of model-internal variations. Such deviations of the latent representation from its mean can be analyzed in the VAE branch:

\vspace{-0.4em}

\subsubsection{VAE branch}
We use VAEs to inspect deviations of the latent representation from its mean. Here, we use the encoders $f_{\mu}$, $f_{\sigma}$, a decoder $g$, and a standard diagonal Gaussian prior $p(z)$, resulting in the VAE objective
\begin{equation}
\label{eq:cevae_objective}
L_{VAE} = L_{KL}(f_{\mu}(x), f_{\sigma}(x)^2) + L_{rec_{VAE}}(x, g(z)), 
\end{equation}
\noindent
where $z \sim \mathcal{N}(f_{\mu}(x) , f_{\sigma}(x)^2 )$ using the reparametrization trick and $L_{KL}$ is the Kullback-Leibler divergence loss (KL-loss) with a standard Gaussian as in Eq. (\ref{eq:vae}).
This density estimation of VAEs is designed to yield a comparable per-sample-likelihood estimate and thus a comparable anomaly score.  
To analyze the deviations of the posterior from the prior of the latent variable distributions we use the KL-loss $L_{KL}$. Below, we show how to trace back these deviations to the pixel level to complement the reconstruction-based delineation of anomalous parts in a data sample.

\vspace{-0.4em}
\subsubsection{ceVAE}
By combining CEs and VAEs, we aim at capturing both effects, namely a better-calibrated reconstruction error and model-internal variations, to yield more complete estimates of anomaly, for each data sample $x$ as well as for the different parts $x_i \in x$ of the sample.
The combined objective function is consequently given as:
\vspace{-0.3em}
\begin{equation}
\label{eq:cevae}
    L_{ceVAE} = L_{KL}(f_{\mu}(x), f_{\sigma}(x)^2) + L_{rec_{VAE}}(x, g(z)) + L_{rec_{CE}}(x, g(f_{\mu}(\tilde{x}))), 
    \vspace{-0.3em}
\end{equation}
where $L_{KL}$ is the KL-loss, $z$ is sampled using the reparametrization trick and $\tilde{x}$ is perturbed by masking out regions as in CEs. 
During training, the CE objective $L_{rec_{CE}}$ does not put constraints for normality on the prior belief $p(z|x)$. This is essential to  prevent the model from deeming such perturbed cases as `normal'.
Furthermore, the combination of a CE and VAE can have a regularizing effect, prevent posterior collapse of the VAE and, due to the CE, lead to representations which capture the semantics of the data better \cite{pathak_context_2016}.

\vspace{-0.5em}

\subsubsection{Anomaly detection}
We can use the ceVAE to detect anomalies on sample and pixel level.
After maximizing the ELBO similar to a VAE, we can estimate the probability $p(x)$ of a data sample $x$ by evaluating the ELBO for a data sample which can give a well-calibrated anomaly score. 
Thus the sample-wise anomaly score is given as:
\vspace{-1.0em}
\begin{equation}
    \log p(x) \approx L_{KL}(x) + L_{rec_{VAE}}(x, g(z)),
    \vspace{-0.5em}
\end{equation}
Simultaneously, to localize abnormal parts in the data sample we combine the reconstruction-based and density-based pixel-wise anomaly scores. 
The recon\-struction-based score is given by the reconstruction error which, due to the denoising task, is geared towards the derivative of the log-density with respect to the input. 
The density-based score is given by a pixel-wise back-tracing of the latent variable deviations from the prior, which is calculated by back propagating the approximated ELBO onto the input. 
This combination results in a more complete estimate of $\frac{\partial \log p(x)}{\partial x}$, thus outlining the ``direction towards normality" for each pixel.
Using an element-wise function $h$ to combine the scores e.g. pixel-wise multiplication, the pixel-wise anomaly score is defined as:
\begin{equation}
\label{eq:pixelscore}
    h \bigg( | x - g(f(x)) |, |\frac{\partial (L_{KL}(x) + L_{rec_{VAE}}(x, z))}{\partial x}| \bigg),
\end{equation}
where the reconstruction error is the absolute pixel-wise difference, and the pixel-wise derivative is calculated by backpropagating the ELBO back onto the data sample.

 \section{Experiments}
 
\vspace{-0.5em}
\subsubsection{Data}

We used T2-weighted images from three different brain MRI datasets. 
The model was trained on the HCP dataset \cite{van_essen_human_2012} to learn the distribution of healthy patients.
After training, the model was tested to detect anomalies in the BraTS-2017 %
\cite{bakas_advancing_2017}
and the ISLES-2015 \cite{maier_isles_2017} dataset.
The HCP dataset, the only dataset used for training, was split into 1092 patients for training and 20 for validation, i.e. 136576 and 2496 slices respectively.
The BraTS-2017 dataset was split into 20 patients for validation and 266 for testing and the ISLES-2015 dataset was split into 8 patients for validation and 20 for testing.
Each dataset was preprocessed similarly, with a patient-wise z-score normalization and slice-wise resampling to a resolution of $64\times64$.
During training, we used random mirroring, rotations, and multiplicative brightness augmentations and the validation data was used to prevent overfitting and choose the best performing model for testing. 
\vspace{-0.5em}
 
\subsubsection{Model}

For the encoder and decoder networks, we chose fully convolutional networks with five 2D-Conv-Layers and 2D-Transposed-Conv-Layers respectively with CoordConv \cite{liu_intriguing_2018}, kernel size 4 and stride 2, each layer followed by a Leaky\-ReLU non-linearity. 
The encoder and decoder are symmetric with 16, 64, 256, 1024 feature maps and a latent variable size of 1024. %
Similar to Kingma et al. \cite{kingma_auto-encoding_2013} the encoders have shared weights, with the last layer having two heads, one predicting the mean, and one predicting the log standard-deviation.
Since it showed similar performance and produced visually slightly sharper images, we chose the L1-Loss instead of the MSE/L2-Loss as reconstruction loss $L_{rec}$. 
Due to different value-ranges of the reconstruction error and the back-propagated values, the combination function $h$ was chosen as element-wise multiplication. To calculate the gradients we used the smoothed guided-backproagation algorithm \cite{smilkov_smoothgrad:_2017,springenberg_striving_2015} and smoothed the gradient with a Gaussian kernel before multiplication because of checkerboard artifacts caused by the Conv-Layers \cite{odena_deconvolution_2016}. Since backpropagating the $L_{rec_{VAE}}$ showed no additional benefit
and only slowed down gradient calculation, we only backpropagate the KL-Loss $L_{KL}$ to the image. 
For CE noise we chose 1-3 randomly sized and positioned squares, but in contrast to Pathak et al. \cite{pathak_context_2016} we chose a random value from the data distribution. This makes the challenge of correcting the noise slightly harder and is conceptually more akin to DAEs with Gaussian noise.
We used Adam with a learning rate of $2 \times 10^{-4}$ and trained the model with a batch size of 64 for 60 epochs. 
\vspace{-0.5em}

\subsubsection{Benchmark Methods}

We compare the proposed model with an OC-SVM and different AE-based methods, which have shown state-of-the-art performance on similar tasks  \cite{baur_deep_2018,chen_unsupervised_2018,chen_deep_2018,pawlowski_unsupervised_2018}. 
The OC-SVM was based on the libsvm implementation \cite{chang_libsvm:_2011}.
For the AE-based methods, we used a standard AE, a DAE, a CE, and a VAE, all using the same model structure and training scheme as the ceVAE.
To further inspect the benefits of combining the CE and VAE, we introduced a ceVAE weighting factor, termed ceVAE-Factor, which indicates the ratio of the CE Loss ($L_{rec_{CE}}$ in Eq. (\ref{eq:cevae})) to the VAE-Loss ($L_{KL}$ and $L_{rec_{VAE}}$ in Eq. (\ref{eq:cevae})). A ratio of $0.0$ implies that the model was trained as a VAE only, a ratio of $1.0$ implies that the model was trained as a CE only, and the other ratios are differently weighted ceVAE models.
\vspace{-0.5em}

\subsubsection{Evaluation-Metrics}
\label{ssec:evaluation}
We separately evaluated the slice/sample-wise performance and the pixel-wise performance.
For the slice-wise evaluation we divided each patient into normal and abnormal slices, depending on the presence of annotations in the slice. Using the estimated sample probability $p(x)$, we evaluated the algorithm on the task to discriminate between normal and abnormal slices and report the ROC-AUC.
For the pixel-wise evaluation, using the pixel-wise anomaly score given by $h$, we determined the pixel-wise ROC-AUC and the mean patient-wise Dice score.
The Dice score is calculated with a 5-fold cross-validation, where we use $\frac{1}{5}$ of the patient-samples to determine an anomaly threshold and apply it on other data samples to determine a segmentation and calculate the mean of a patient-wise Dice score. 
As anomaly-labels the ground-truth annotations were used, considering all annotations as anomalies. 
For each model, we performed five runs and report the median as well as the max and min performance.
\vspace{-0.8em}

\section{Results}
\vspace{-0.7em}

Given the proposed framework we first evaluated the effect of combining a CE with a VAE for slice-wise anomaly detection. This is followed by an evaluation of the benefits of combining the reconstruction error with the gradient of the KL-Loss for a pixel-wise detection.
\vspace{-1.0em}

\subsubsection{Slice-wise detection}

\begin{figure}[tb]
  \begin{minipage}[c]{0.65\textwidth}
    \includegraphics[width=0.99\textwidth]{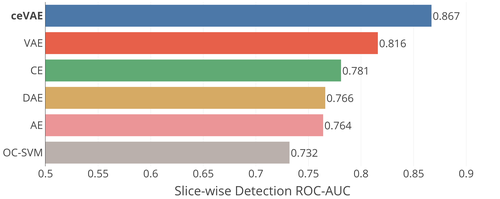}
  \end{minipage}\hfill
  \begin{minipage}[c]{0.33\textwidth}
    \caption{Comparison of slice-wise anomaly detection performance of different models on the BraTS-2017 dataset.}
   \label{plt:slicewise}
  \end{minipage}
\vspace{-1.0em}
\end{figure}
\vspace{-0.1em}

Firstly we compared the performance of different approaches on the slice-wise anomaly detection task. 
Fig. \ref{plt:slicewise} shows the performance of different methods on the BraTS 2017 dataset. As often reported, the OC-SVM had difficulties with the structured and high-dimensional data \cite{goldstein_comparative_2016}. 
An AE outperformed the OC-SVM on this task, and could further be improved upon by using an auxiliary denoising task, where context encoding appeared to be more fitting in this case. 
Using a standard VAE could further improve the performance, while the ceVAE outperformed all other methods by a margin.
\vspace{-1.0em}

\subsubsection{Pixel-wise detection}

\begin{table}[tb]

\begin{minipage}[c]{0.025\linewidth}
\centering
\includegraphics[width=\linewidth]{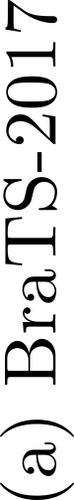}
\end{minipage}
\hspace{0.1cm}
\begin{minipage}[c]{0.96\linewidth}
    \centering
    \begin{subfigure}[b]{0.48\textwidth}
               \includegraphics[width=\textwidth]{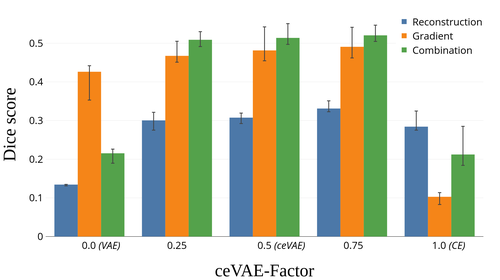}
        \end{subfigure}
        \hspace{0.1cm}
        \begin{subfigure}[b]{0.48\textwidth}
               \includegraphics[width=\textwidth]{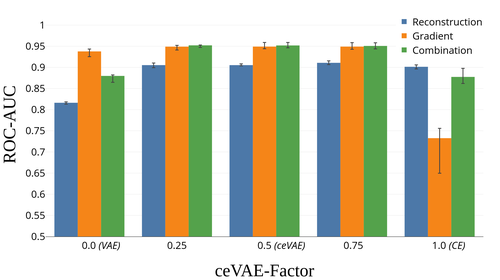}
        \end{subfigure}
\end{minipage}

\vspace{0.7em}

\begin{minipage}[c]{0.025\linewidth}
\centering
\includegraphics[width=\linewidth]{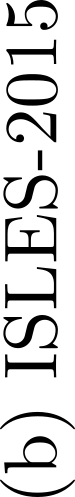}
\end{minipage}
\hspace{0.1cm}
\begin{minipage}[c]{0.96\linewidth}
    \centering
    \begin{subfigure}[b]{0.48\textwidth}
        \includegraphics[width=\textwidth]{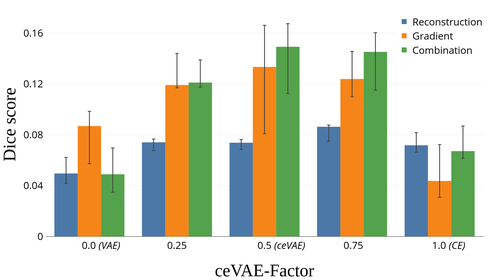}
    \end{subfigure}        
    \hspace{0.1cm}
    \begin{subfigure}[b]{0.48\textwidth}
        \includegraphics[width=\textwidth]{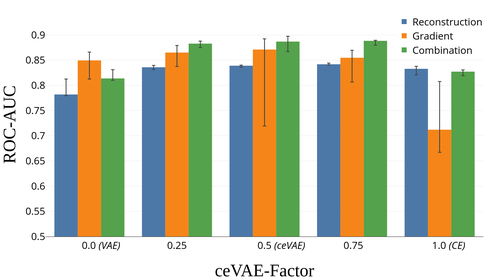}
    \end{subfigure}
    
\end{minipage}

\vspace{0.7em}
\captionof{figure}{Dice score and pixel-wise ROC-AUC for different ceVAE-Factors. 
}
\label{plt:pixelwise}
\vspace{-2.0em}
\end{table}

For the pixel-wise performance we focused on CE, VAE, and ceVAE, since these were the best performing models in the slice-wise task and since VAEs have become a de-facto standard in anomaly-detection for images 
\cite{baur_deep_2018,chen_deep_2018,kiran_overview_2018}.
We report the pixel-wise ROC-AUC and Dice scores on the BraTS-2017 and ISLES-2015 datasets in Fig. \ref{plt:pixelwise}.

Results for the non-combined methods were as expected: the CE performed best when using solely the reconstruction error (first argument of $h$, Eq. (\ref{eq:pixelscore})), while the VAE performed best when using solely the gradient of the KL-loss (second argument of $h$, Eq. (\ref{eq:pixelscore})), outperforming the CE.
The ceVAE (combination of CE and VAE) outperformed the non-combined methods in all cases, while a combination of the reconstruction error and the KL-loss gradient yielded the best results throughout the experiments.
Focusing on the reconstruction error only, it is interesting to note that a combination of a VAE with a CE already shows benefits, possibly due to the regularizing effects described in Sec \ref{sssec:ceVAE}.
It is also important to notice the difference in absolute performance on the different datasets.
One probable explanation is the difference in dataset quality and thus the data distribution to start with.
For each dataset, we show some qualitative results in Fig. \ref{fig:qualitative}.

\begin{table}[tb]

\begin{minipage}[c]{0.07\linewidth}
BraTS-2017

\end{minipage}
\hspace{0.1cm}
\begin{minipage}[c]{0.91\linewidth}
    \centering
    \begin{subfigure}{0.02\textwidth}
        \includegraphics[width=\textwidth]{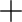}
    \end{subfigure}
    \begin{subfigure}{0.46\textwidth}
        \includegraphics[width=\textwidth]{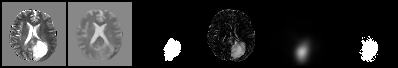}
    \end{subfigure}
    \begin{subfigure}{0.46\textwidth}
        \includegraphics[width=\textwidth]{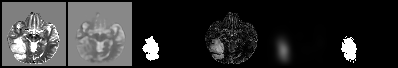}
    \end{subfigure}
    \begin{subfigure}{0.02\textwidth}
        \includegraphics[width=\textwidth]{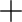}
    \end{subfigure}
    
    \begin{subfigure}{0.02\textwidth}
        \includegraphics[width=\textwidth]{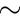}
    \end{subfigure}
    \begin{subfigure}{0.46\textwidth}
        \includegraphics[width=\textwidth]{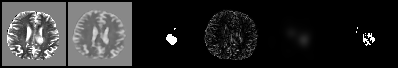}
    \end{subfigure}
    \begin{subfigure}{0.46\textwidth}
        \includegraphics[width=\textwidth]{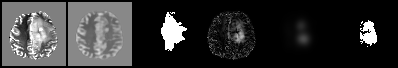}
    \end{subfigure}
    \begin{subfigure}{0.02\textwidth}
        \includegraphics[width=\textwidth]{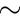}
    \end{subfigure}
    
    \begin{subfigure}{0.02\textwidth}
        \includegraphics[width=\textwidth]{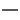}
    \end{subfigure}
    \begin{subfigure}{0.46\textwidth}
        \includegraphics[width=\textwidth]{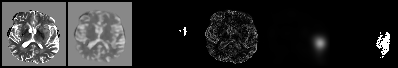}
    \end{subfigure}
    \begin{subfigure}{0.46\textwidth}
        \includegraphics[width=\textwidth]{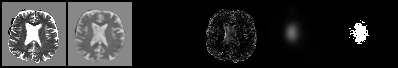}
    \end{subfigure}
    \begin{subfigure}{0.02\textwidth}
        \includegraphics[width=\textwidth]{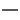}
    \end{subfigure}
\end{minipage}

\hrule

\begin{minipage}[c]{0.07\linewidth}
ISLES-2015
\end{minipage}
\hspace{0.1cm}
\begin{minipage}[c]{0.91\linewidth}
    \centering
    \begin{subfigure}{0.02\textwidth}
        \includegraphics[width=\textwidth]{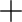}
    \end{subfigure}
    \begin{subfigure}{0.46\textwidth}
        \includegraphics[width=\textwidth]{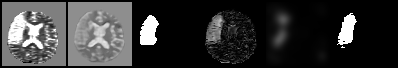}
    \end{subfigure}
    \begin{subfigure}{0.46\textwidth}
        \includegraphics[width=\textwidth]{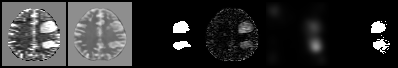}
    \end{subfigure}
    \begin{subfigure}{0.02\textwidth}
        \includegraphics[width=\textwidth]{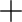}
    \end{subfigure}
    
    \begin{subfigure}{0.02\textwidth}
        \includegraphics[width=\textwidth]{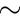}
    \end{subfigure}
    \begin{subfigure}{0.46\textwidth}
        \includegraphics[width=\textwidth]{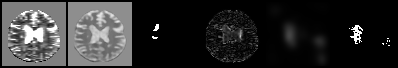}
    \end{subfigure}
    \begin{subfigure}{0.46\textwidth}
        \includegraphics[width=\textwidth]{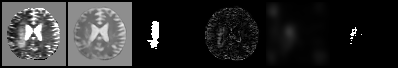}
    \end{subfigure}
    \begin{subfigure}{0.02\textwidth}
        \includegraphics[width=\textwidth]{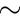}
    \end{subfigure}
    
    \begin{subfigure}{0.02\textwidth}
        \includegraphics[width=\textwidth]{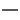}
    \end{subfigure}
    \begin{subfigure}{0.46\textwidth}
        \includegraphics[width=\textwidth]{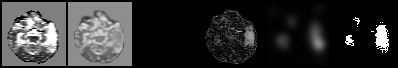}
    \end{subfigure}
    \begin{subfigure}{0.46\textwidth}
        \includegraphics[width=\textwidth]{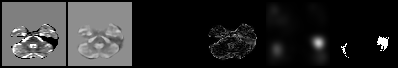}
    \end{subfigure}
    \begin{subfigure}{0.02\textwidth}
        \includegraphics[width=\textwidth]{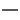}
    \end{subfigure}
    
\end{minipage}

\hrule
\begin{minipage}[c]{0.07\linewidth}
\centering
HCP
\end{minipage}
\hspace{0.1cm}
\begin{minipage}[c]{0.91\linewidth}
    \centering
    \begin{subfigure}{0.02\textwidth}
        \includegraphics[width=\textwidth]{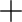}
    \end{subfigure}
    \begin{subfigure}{0.46\textwidth}
        \includegraphics[width=\textwidth]{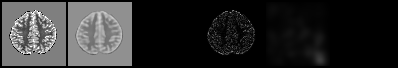}
    \end{subfigure}
    \begin{subfigure}{0.46\textwidth}
        \includegraphics[width=\textwidth]{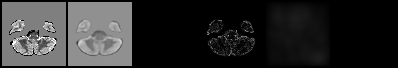}
    \end{subfigure}
    \begin{subfigure}{0.02\textwidth}
        \includegraphics[width=\textwidth]{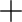}
    \end{subfigure}
    
    \begin{subfigure}{0.02\textwidth}
        \includegraphics[width=\textwidth]{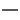}
    \end{subfigure}
    \begin{subfigure}{0.46\textwidth}
        \includegraphics[width=\textwidth]{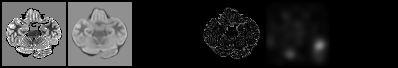}
    \end{subfigure}
    \begin{subfigure}{0.46\textwidth}
        \includegraphics[width=\textwidth]{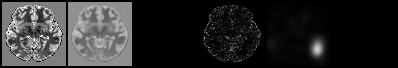}
    \end{subfigure}
    \begin{subfigure}{0.02\textwidth}
        \includegraphics[width=\textwidth]{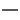}
    \end{subfigure}

\end{minipage}

\begin{minipage}[c]{0.07\linewidth}
\centering
\includegraphics[width=0.3\textwidth]{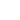}
\end{minipage}
\hspace{0.1cm}
\begin{minipage}[c]{0.91\linewidth}
    \centering
    \begin{subfigure}{0.02\textwidth}
        \includegraphics[width=\textwidth]{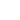}
    \end{subfigure}
    \begin{subfigure}{0.46\textwidth}
        \includegraphics[width=\textwidth]{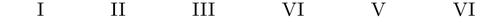}
    \end{subfigure}
    \begin{subfigure}{0.46\textwidth}
        \includegraphics[width=\textwidth]{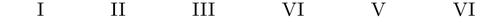}
    \end{subfigure}
    \begin{subfigure}{0.02\textwidth}
        \includegraphics[width=\textwidth]{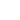}
    \end{subfigure}
    
\end{minipage}

\vspace{0.7em}
\captionof{figure}{Sample images from test-sets. The 1st, 2nd, and 3rd row show good (+), medium ($\sim$), and failure (-) cases respectively. For each sample the original sample (I), the reconstruction (II), the annotation (III), the reconstruction error (IV), the gradient (V), and the resulting segmentation (VI) are presented.
}
\label{fig:qualitative}
\vspace{-3.0em}
\end{table}

\vspace{-0.5em}

\section{Discussion \& Conclusion}
\vspace{-0.5em}

In this work we present ceVAE for unsupervised anomaly detection, combining CEs with VAEs for unsupervised training and detection as well as localization of anomalies in medical images. We demonstrate the performance gain over the individual approaches and outperform all presented baselines as well as the results in the literature \cite{chen_unsupervised_2018,chen_deep_2018}. We further show how the approach can be used for a pixel-wise localization of the anomalies, achieving state-of-the-art ROC-AUCs for unsupervised segmentation on public benchmark data. 

Evaluating the performance of an anomaly detection algorithm is a challenging problem. Since there is no reference anomaly detection dataset in the field, surrogate datasets are used. Not all anomalies in the dataset might be labeled, thus the performance on those datasets might lower bound the actual performance.
The domain shifts between different datasets can also obstruct the evaluation. 
In the HCP training dataset, the patients are healthy with an age of 25-35 years, all recorded on the same scanner type with a high spatial resolution. In contrast, in the BraTS-2017 and ISLES-2015 test datasets, most patients are older and different scanners across multiple institutions with varying image quality were used. This results in two additional distribution shifts, age and image quality, which can cause additional miss-detections. This is especially evident in the ISLES-2015 dataset, where the image quality is quite low, potentially explaining the low absolute scores in the results.

Despite these challenges, the proposed approach yields relatively strong results for unsupervised segmentation and outperforms other state-of-the-art methods on the given datasets \cite{chen_unsupervised_2018,chen_deep_2018}.
We evaluated different parameter settings and design choices. Adding more layers, residual connections, different normalization-layers, and/or using pixel reshuffling %
as downsampling operation did not yield any significant benefits, and thus for the sake of Occam's-Razor and training speed, we chose to keep our simple (``first educated guess") model. 
Using 2.5D input, i.e. using some previous and consecutive slides did not show any significant benefits either, thus we did not include it in the final models, but extending the work to 3D might be an interesting next step. Early results on a resolution of $128\times128$ and $192\times192$ pixels showed a similar or slightly better performance (a full analysis is currently omitted due to time constraints).

Despite the results discussed by Adebayo et al. \cite{adebayo_sanity_2018} we could not find any model or output independence of the guided backpropagation algorithm, and it slightly outperformed vanilla backpropagation. 
We also tried replacing/augmenting the KL-Loss with an MMD-Loss 
or an Adversarial-Loss, 
which were reported to slightly boost the performance \cite{chen_unsupervised_2018}, but while showing a minor boost in reconstruction error, due to higher variance gradients the overall performance deteriorated.
Using different reconstruction losses, such as MSE, an Adversarial or Feature-Loss, %
despite making the reconstructions less blurry, did not show any significant performance benefits and were omitted due to their increased training time and unstable training regime. It might be an interesting future direction to see how different (perceptual) reconstruction losses
can further boost the performance or interpretability. %
Another future direction of research might be to integrate sampling into the anomaly score estimation. Using a bigger sampling size 
for the MC-sampling of the VAE might give insights into areas where the learned data distribution is not well represented and thus indicates anomalies. Similarly dropout sampling might be an alternative and could further aid the performance as well.

We have presented a combination of a density-based and reconstruction-based anomaly detection approaches, which does not need labeled data and also allows for a sample-wise anomaly scoring and localization of the anomalies. The results are promising and have the potential to improve and speed up the future inspection and evaluation of medical images, thus supporting physicians in coping with the increasing amounts of medical imaging data being produced.

\bibliographystyle{splncs04}
\bibliography{refs/refs_short4}

\begin{thebibliography}{10}
\providecommand{\url}[1]{\texttt{#1}}
\providecommand{\urlprefix}{URL }
\providecommand{\doi}[1]{https://doi.org/#1}

\bibitem{adebayo_sanity_2018}
Adebayo, J., Gilmer, J., Muelly, M., Goodfellow, I., Hardt, M., Kim, B.: Sanity
  {Checks} for {Saliency} {Maps}  (2018)

\bibitem{alain_what_2014}
Alain, G., Bengio, Y.: What {Regularized} {Auto}-encoders {Learn} from the
  {Data}-generating {Distribution}. JMLR  (2014)

\bibitem{an_variational_2015}
An, J., Cho, S.: Variational {Autoencoder} based {Anomaly} {Detection} using
  {Reconstruction} {Probability} (2015)

\bibitem{bach_breaking_2017}
Bach, F.R.: Breaking the {Curse} of {Dimensionality} with {Convex} {Neural}
  {Networks}. JMLR  (2017)

\bibitem{bakas_advancing_2017}
Bakas, S., Akbari, H., Sotiras, A., Bilello, M., Rozycki, M., Kirby, J.S.,
  Freymann, J.B., Farahani, K., Davatzikos, C.: Advancing {The} {Cancer}
  {Genome} {Atlas} glioma {MRI} collections with expert segmentation labels and
  radiomic features. Sci Data  (2017)

\bibitem{ballard_modular_1987}
Ballard, D.H.: Modular {Learning} in {Neural} {Networks}. In: {AAAI} (1987)

\bibitem{baur_deep_2018}
Baur, C., Wiestler, B., Albarqouni, S., Navab, N.: Deep {Autoencoding} {Models}
  for {Unsupervised} {Anomaly} {Segmentation} in {Brain} {MR} {Images}. CoRR
  (2018)

\bibitem{bluemke_chapter_2012}
Bluemke, D.A., Liu, S.: Chapter 41 - {Imaging} in {Clinical} {Trials}. In:
  Principles and {Practice} of {Clinical} {Research} ({Third} {Edition}).
  Academic Press (2012)

\bibitem{chang_libsvm:_2011}
Chang, C.C., Lin, C.J.: {LIBSVM}: {A} library for support vector machines. ACM
  TIST  (2011)

\bibitem{chen_unsupervised_2018}
Chen, X., Konukoglu, E.: Unsupervised {Detection} of {Lesions} in {Brain} {MRI}
  using constrained adversarial auto-encoders. CoRR  (2018)

\bibitem{chen_deep_2018}
Chen, X., Pawlowski, N., Rajchl, M., Glocker, B., Konukoglu, E.: Deep
  {Generative} {Models} in the {Real}-{World}: {An} {Open} {Challenge} from
  {Medical} {Imaging}. CoRR  (2018)

\bibitem{drew_invisible_2013}
Drew, T., Vo, M.L.H., Wolfe, J.M.: “{The} invisible gorilla strikes again:
  {Sustained} inattentional blindness in expert observers”. Psychol Sci
  (2013)

\bibitem{goldstein_comparative_2016}
Goldstein, M., Uchida, S.: A {Comparative} {Evaluation} of {Unsupervised}
  {Anomaly} {Detection} {Algorithms} for {Multivariate} {Data}. PLoS ONE
  (2016)

\bibitem{kingma_auto-encoding_2013}
Kingma, D.P., Welling, M.: Auto-{Encoding} {Variational} {Bayes}. CoRR  (2013)

\bibitem{kiran_overview_2018}
Kiran, B., Thomas, D., Parakkal, R., Kiran, B.R., Thomas, D.M., Parakkal, R.:
  An {Overview} of {Deep} {Learning} {Based} {Methods} for {Unsupervised} and
  {Semi}-{Supervised} {Anomaly} {Detection} in {Videos}. Journal of Imaging
  (2018)

\bibitem{liu_intriguing_2018}
Liu, R., Lehman, J., Molino, P., Such, F.P., Frank, E., Sergeev, A., Yosinski,
  J.: An {Intriguing} {Failing} of {Convolutional} {Neural} {Networks} and the
  {CoordConv} {Solution}. CoRR  (2018)

\bibitem{maier_isles_2017}
Maier, O., Menze, B.H., von~der Gablentz, J., Hani, L., Heinrich, M.P.,
  Liebrand, M., Winzeck, S., Basit, A., Bentley, P., Chen, L., Christiaens, D.,
  Dutil, F., Egger, K., Feng, C., Glocker, B., Götz, M., Haeck, T., Halme,
  H.L., Havaei, M., Iftekharuddin, K.M., Jodoin, P.M., Kamnitsas, K., Kellner,
  E., Korvenoja, A., Larochelle, H., Ledig, C., Lee, J.H., Maes, F., Mahmood,
  Q., Maier-Hein, K.H., McKinley, R., Muschelli, J., Pal, C., Pei, L.,
  Rangarajan, J.R., Reza, S.M.S., Robben, D., Rueckert, D., Salli, E., Suetens,
  P., Wang, C.W., Wilms, M., Kirschke, J.S., Kr~Amer, U.M., Münte, T.F.,
  Schramm, P., Wiest, R., Handels, H., Reyes, M.: {ISLES} 2015 - {A} public
  evaluation benchmark for ischemic stroke lesion segmentation from
  multispectral {MRI}. Med Image Anal  (2017)

\bibitem{odena_deconvolution_2016}
Odena, A., Dumoulin, V., Olah, C.: Deconvolution and {Checkerboard}
  {Artifacts}. Distill  (2016)

\bibitem{pathak_context_2016}
Pathak, D., Krähenbühl, P., Donahue, J., Darrell, T., Efros, A.A.: Context
  {Encoders}: {Feature} {Learning} by {Inpainting}. CVPR  (2016)

\bibitem{pawlowski_unsupervised_2018}
Pawlowski, N., Lee, M.C.H., Rajchl, M., McDonagh, S., Ferrante, E., Kamnitsas,
  K., Cooke, S., Stevenson, S.K., Khetani, A.M., Newman, T., Zeiler, F.A.,
  Digby, R.J., Coles, J.P., Rueckert, D., Menon, D.K., Newcombe, V.F.J.,
  Glocker, B.: Unsupervised {Lesion} {Detection} in {Brain} {CT} using
  {Bayesian} {Convolutional} {Autoencoders} (2018)

\bibitem{rezende_stochastic_2014}
Rezende, D.J., Mohamed, S., Wierstra, D.: Stochastic {Backpropagation} and
  {Approximate} {Inference} in {Deep} {Generative} {Models}. In: {ICML}.
  JMLR.org (2014)

\bibitem{rifai_contractive_2011}
Rifai, S., Vincent, P., Muller, X., Glorot, X., Bengio, Y.: Contractive
  {Auto}-encoders: {Explicit} {Invariance} {During} {Feature} {Extraction}. In:
  {ICML} (2011)

\bibitem{schlegl_unsupervised_2017}
Schlegl, T., Seeböck, P., Waldstein, S.M., Schmidt-Erfurth, U., Langs, G.:
  Unsupervised {Anomaly} {Detection} with {Generative} {Adversarial} {Networks}
  to {Guide} {Marker} {Discovery}. In: {IPMI}. Springer (2017)

\bibitem{scholkopf_estimating_2001}
Schölkopf, B., Platt, J.C., Shawe-Taylor, J.C., Smola, A.J., Williamson, R.C.:
  Estimating the {Support} of a {High}-{Dimensional} {Distribution}. Neural
  Comput.  (2001)

\bibitem{shyu_novel_2003}
Shyu, M.L., Chen, S.C., Sarinnapakorn, K., Chang, L.: A {Novel} {Anomaly}
  {Detection} {Scheme} {Based} on {Principal} {Component} {Classifier}. ICDM
  (2003)

\bibitem{smilkov_smoothgrad:_2017}
Smilkov, D., Thorat, N., Kim, B., Viégas, F.B., Wattenberg, M.: {SmoothGrad}:
  removing noise by adding noise. CoRR  (2017)

\bibitem{springenberg_striving_2015}
Springenberg, J.T., Dosovitskiy, A., Brox, T., Riedmiller, M.: Striving for
  {Simplicity}: {The} {All} {Convolutional} {Net}. In: {ICLR} (workshop track)
  (2015)

\bibitem{van_essen_human_2012}
Van~Essen, D.C., Ugurbil, K., Auerbach, E., Barch, D., Behrens, T.E.J.,
  Bucholz, R., Chang, A., Chen, L., Corbetta, M., Curtiss, S.W., Della~Penna,
  S., Feinberg, D., Glasser, M.F., Harel, N., Heath, A.C., Larson-Prior, L.,
  Marcus, D., Michalareas, G., Moeller, S., Oostenveld, R., Petersen, S.E.,
  Prior, F., Schlaggar, B.L., Smith, S.M., Snyder, A.Z., Xu, J., Yacoub, E.,
  {WU-Minn HCP Consortium}: The {Human} {Connectome} {Project}: a data
  acquisition perspective. Neuroimage  (2012)

\bibitem{vernooij_incidental_2007}
Vernooij, M.W., Ikram, M.A., Tanghe, H.L., Vincent, A.J., Hofman, A., Krestin,
  G.P., Niessen, W.J., Breteler, M.M., van~der Lugt, A.: Incidental {Findings}
  on {Brain} {MRI} in the {General} {Population}. NEJM  (2007)

\bibitem{vincent_stacked_2010}
Vincent, P., Larochelle, H., Lajoie, I., Bengio, Y., Manzagol, P.A.: Stacked
  {Denoising} {Autoencoders}: {Learning} {Useful} {Representations} in a {Deep}
  {Network} with a {Local} {Denoising} {Criterion}. JMLR  (2010)

\end{thebibliography}

\end{document}